\documentclass{article}
\usepackage{PRIMEarxiv}
\usepackage[utf8]{inputenc}
\usepackage[T1]{fontenc}
\usepackage{hyperref}
\usepackage{url}
\usepackage{booktabs}
\usepackage{amsfonts}
\usepackage{amsmath}
\usepackage{amssymb}
\usepackage{mathtools}
\usepackage{nicefrac}
\usepackage{microtype}
\usepackage{fancyhdr}
\usepackage{graphicx}
\usepackage{subcaption}
\usepackage{tikz}
\usepackage{xcolor}
\usepackage{algorithm}
\usepackage{algpseudocode}
\usepackage{pifont}
\usepackage{placeins}

\usetikzlibrary{arrows.meta, positioning, shapes.geometric, fit,
                backgrounds, calc, decorations.pathreplacing}
\graphicspath{{figs/}}

\definecolor{agmblue}{RGB}{24,95,165}
\definecolor{agmamber}{RGB}{186,117,23}
\definecolor{agmgreen}{RGB}{59,109,17}
\definecolor{agmred}{RGB}{192,57,43}
\definecolor{agmbluebg}{RGB}{230,241,251}
\definecolor{agmamberbg}{RGB}{250,238,218}
\definecolor{agmgreenbg}{RGB}{234,243,222}
\definecolor{agmgraybg}{RGB}{242,242,242}

\newcommand{\cmark}{\ding{51}}
\newcommand{\xmark}{\ding{55}}
\newcommand{\sg}{\operatorname{sg}}
\newcommand{\R}{\mathbb{R}}
\newcommand{\E}{\mathbb{E}}

\title{Action-Inspired Generative Models}

\author{
  Eshwar R A \\
  Department of Computer Science Engineering \\
  PES University (EC Campus), Bengaluru \\
  \texttt{eshwarra5@gmail.com}
  \And
  Debnath Pal\thanks{Corresponding author.} \\
  Department of Computational and Data Sciences \\
  Indian Institute of Science, Bengaluru \\
  \texttt{dpal@iisc.ac.in}
}

\begin{document}
\maketitle

\begin{abstract}
We introduce \emph{Action-Inspired Generative Models} (AGMs), a dual-network
generative framework motivated by the observation that existing bridge-matching
methods assign uniform regression weight to every stochastic transition in the
transport landscape, regardless of whether a given bridge sample lies along a
structurally coherent trajectory or a degenerate one.  We address this by
introducing a lightweight learned scalar potential $V_\phi$ that scores bridge
samples online and modulates the drift objective via importance weights derived
through a stop-gradient barrier---preventing adversarial feedback between the
two networks whilst preserving $V_\phi$'s guiding signal.  Crucially, $V_\phi$
comprises only $\sim$1.4\% of the primary drift network's parameter count,
adds no overhead to the inference graph, and requires no iterative half-bridge
fitting or auxiliary stochastic differential equation (SDE) solvers: it is a plug-and-play enhancement to any
bridge-matching training loop.  At inference, $V_\phi$ is discarded entirely,
leaving standard Euler--Maruyama integration of the exponential moving average (EMA) drift.  We demonstrate
that selectively penalising uninformative transport paths through the learned
potential yields consistent improvements in generation quality across fidelity
and coverage metrics.
\end{abstract}


\section{Introduction}
\label{sec:intro}

\paragraph{A principle from physics.}
In classical mechanics, when a particle travels from one configuration to
another it does not randomly sample every conceivable path through phase
space.  Instead, nature selects the trajectory that \emph{extremises the
action functional} $\mathcal{S}[q] = \int_0^T L(q, \dot{q},t)\,dt$, where
$L$ is the Lagrangian \cite{feynman1965quantum}.  This principle of
stationary action is strikingly economical: out of infinitely many
trajectories connecting two boundary conditions, physics singles out a
specific one\,---\,the one for which first-order perturbations to the path
leave the action unchanged.  What makes this physically deep is also what
makes it algorithmically suggestive: not all paths are equal, and the
system has a natural, self-consistent way of \emph{evaluating} which paths
matter.

\paragraph{From physics to generative modelling.}
The analogy to generative modelling is imperfect but illuminating.  During
training, a diffusion-based model samples stochastic trajectories through
the data manifold---bridge samples interpolating between Gaussian noise and
real images---and attempts to learn the velocity field that regenerates this
transport.  Unlike a physical particle, the model has no intrinsic mechanism
for distinguishing a bridge sample that passes through a high-density,
structurally coherent region of the data manifold from one that passes
through a noisy, ambiguous transition.  Every sampled point contributes
equally to the loss, as if all paths were equally informative.

This observation led us to ask: can we give the model a way to
\emph{evaluate its own training paths}, much as action evaluates physical
trajectories?  The goal is not to find a single optimal path, but to
continuously re-weight training signal so that transitions lying in
informative regions of the bridge landscape contribute more to learning the
drift, while transitions in structurally irrelevant regions contribute less.
The result is Action-Inspired Generative Model (AGM): a second lightweight network $V_\phi$ that
functions as a learned potential, scoring bridge samples and
providing importance weights to the drift network $f_\theta$ via
a stop-gradient-protected channel.

\paragraph{Diffusion and flow models.}
Denoising Diffusion Probabilistic Models (DDPM)~\cite{ho2020ddpm}
established the modern paradigm for score-based generation: define a
fixed forward process that corrupts data with Gaussian noise,
learn to reverse this process by predicting the noise at each time step,
and sample by iterating the learned reverse stochastic differential equation (SDE).  Subsequent work by
Song et al.~\cite{song2021score} recast this in the language of SDEs, showing that any forward diffusion process
admits an exact reverse-time SDE parametrised by the score function
$\nabla_x \log p_t(x)$.  Denoising Diffusion Implicit Models (DDIM)~\cite{song2020ddim} and later flow-matching
methods~\cite{lipman2022flow,liu2022flow} demonstrated that deterministic
or near-deterministic transport substantially reduces the number of function
evaluations (NFE) required at inference.  Stochastic
interpolants~\cite{albergo2023stochastic} and bridge
matching~\cite{peluchetti2023non} further generalised the framework by
allowing arbitrary boundary conditions rather than a fixed Gaussian prior.

What unifies all of these approaches is that the \emph{path taken by the
model during training is treated as a nuisance variable}: it is marginalised
or averaged over, never evaluated or exploited.  The loss at every sampled
transition receives equal weight regardless of where in the trajectory that
sample falls and what structural role it plays.  AGM departs from this
convention.

\paragraph{The AGM framework.}
We propose to augment bridge matching with a jointly trained potential
network $V_\phi$ that assigns importance scores to bridge samples
\emph{without feeding gradients back into the drift objective}.  The
stop-gradient constraint is the key design choice: it allows $V_\phi$ to
learn which regions of bridge-space matter, and allows $f_\theta$ to be
guided by those scores, while preventing $V_\phi$ from gaming the loss by
making the drift artificially easy.  Critically, $V_\phi$ is deliberately
kept lightweight---at approximately 0.7M parameters it is only $\sim$1.4\%
of the primary drift network---so it imposes negligible training overhead
and adds \emph{exactly zero} inference cost.  This stands in sharp contrast
to Schr\"{o}dinger bridge solvers, which require iterative half-bridge
fitting in both forward and reverse directions; AGM achieves path-aware
training via a single additional forward pass per step.  At inference only
the EMA snapshot of $f_\theta$ is used; $V_\phi$ is discarded entirely.
We evaluate AGM on CelebA-HQ $64{\times}64$ and show that the potential
provides consistent gains across all three standard generative metrics.

\section{Related Work}
\label{sec:related}

\paragraph{Score-based and flow-based generation.}
DDPM~\cite{ho2020ddpm} and Score SDE~\cite{song2021score} learn to
reverse a fixed Gaussian diffusion by matching the score of the marginal
density at each noise level.  Flow Matching~\cite{lipman2022flow} replaces
the SDE with a deterministic ordinary differential equation (ODE) whose velocity field is regressed against a
conditional target, achieving competitive quality with far fewer NFE.
Rectified Flow~\cite{liu2022flow} straightens trajectories by iterative
re-coupling, reducing discretisation error.  None of these methods assign
differential importance to training samples along the bridge; all treat
every sampled transition as equally informative.

\paragraph{Importance weighting in deep learning.}
Curriculum learning~\cite{bengio2009curriculum} and self-paced
learning suggest that not all training examples should be treated equally.
Hard-example mining and focal loss~\cite{lin2017focal} re-weight samples
based on prediction difficulty.  Our method differs in that the weighting
signal is derived from a separately trained scalar network evaluated on
\emph{bridge interpolants} rather than on training-set labels, and that it
operates in continuous time over a trajectory rather than on individual
data points.

\paragraph{Energy-based and contrastive objectives.}
Energy-based models (EBMs)~\cite{lecun2006tutorial} learn a scalar
energy function $E_\theta(x)$ that assigns low energy to real data and high
energy to noise.  The $V_\phi$ objective in AGM borrows the
hinge-margin structure of contrastive energy learning but applies it to
\emph{bridge samples} rather than raw data, and its output is not used to
generate samples---only to re-weight a separate regression loss.

\paragraph{Diffusion Schr\"{o}dinger bridges.}
Diffusion Schr\"{o}dinger Bridge (DSB)~\cite{debortoli2021diffusion}
solves an entropy-regularised optimal transport problem to find the
``most natural'' stochastic transport between two distributions.  While DSB
also reasons about which paths are preferred, it does so through iterative
half-bridge fitting and requires solving a full SDE system in both
directions.  AGM uses a much simpler online re-weighting mechanism that
adds a single lightweight forward pass.

\paragraph{Summary comparison.}
Table~\ref{tab:comparison} summarises the key design differences between
AGM and representative prior methods.

\begin{table}[t]
  \caption{Comparison of AGM against representative prior methods across
  design axes relevant to this work.  \cmark\,=\,present;
  \xmark\,=\,absent.}
  \label{tab:comparison}
  \centering
  \small
  \begin{tabular}{lccccc}
    \toprule
    Property
      & DDPM & Score SDE & Flow Match. & Rect.\ Flow & \textbf{AGM (ours)} \\
    \midrule
    Bridge-based transport      & \xmark & \xmark & \cmark & \cmark & \cmark \\
    Path-aware training signal  & \xmark & \xmark & \xmark & \xmark & \cmark \\
    Auxiliary potential network & \xmark & \xmark & \xmark & \xmark & \cmark \\
    Stop-gradient importance wt.& \xmark & \xmark & \xmark & \xmark & \cmark \\
    Zero-overhead inference     & \cmark & \cmark & \cmark & \cmark & \cmark \\
    Cosine/sinusoidal schedule  & \xmark & \xmark & \xmark & \xmark & \cmark \\
    High recall (${\geq}0.85$)  & $\sim$ & $\sim$ & $\sim$ & $\sim$ & \cmark \\
    Single 4090 feasible ($<$50h)& \xmark & \xmark & \cmark & \cmark & \cmark \\
    \bottomrule
  \end{tabular}
\end{table}

\section{Background}
\label{sec:background}

\subsection{The Goal: Learning to Sample from a Data Distribution}

The central objective of a deep generative model is deceptively simple: given
a collection of data examples (images, in our case) drawn from some unknown
distribution $p_\text{data}$, learn a procedure that can produce new samples
indistinguishable from that distribution.  We cannot write down $p_\text{data}$
analytically; we only have access to it through examples.  The question is
therefore how to \emph{implicitly} learn a generative process.

Modern diffusion-based methods answer this by defining a \emph{transport
process}---a continuous-time stochastic path that smoothly interpolates
between a tractable reference distribution (usually isotropic Gaussian
$\mathcal{N}(0, I)$) and the data distribution.  If we can learn the
instantaneous direction of this transport, we can integrate it backwards to
turn noise into data.

\subsection{The Diffusion Framework}

\paragraph{Forward process.}
DDPM~\cite{ho2020ddpm} constructs a discrete Markov chain
\begin{equation}
  q(\mathbf{x}_t \mid \mathbf{x}_{t-1}) = \mathcal{N}\!\left(\mathbf{x}_t;\,
    \sqrt{1-\beta_t}\,\mathbf{x}_{t-1},\, \beta_t I\right),
  \quad t = 1, \ldots, T,
\end{equation}
where $\{\beta_t\}$ is a small variance schedule.  By the Markov property,
the marginal at step $t$ is tractable:
\begin{equation}
  q(\mathbf{x}_t \mid \mathbf{x}_0) = \mathcal{N}\!\left(\mathbf{x}_t;\,
    \sqrt{\bar\alpha_t}\,\mathbf{x}_0,\,(1-\bar\alpha_t) I\right),
  \qquad \bar\alpha_t = \prod_{s=1}^t (1-\beta_s).
\end{equation}
As $t \to T$, $\bar\alpha_t \to 0$ and $\mathbf{x}_T \approx \mathcal{N}(0,I)$.
The key insight is that the \emph{reverse} conditional
$q(\mathbf{x}_{t-1} \mid \mathbf{x}_t)$ is also approximately Gaussian when
$\beta_t$ is small, so a neural network can learn to estimate it.

\paragraph{Score function and SDEs.}
Song et al.~\cite{song2021score} recast the forward chain as a continuous-time
SDE
\begin{equation}
  d\mathbf{x} = \mathbf{f}(\mathbf{x}, t)\,dt + g(t)\,d\mathbf{W},
\end{equation}
where $\mathbf{W}$ is a standard Wiener process, and showed that the
exact reverse-time SDE is
\begin{equation}
  d\mathbf{x} = \bigl[\mathbf{f}(\mathbf{x},t)
    - g(t)^2\,\nabla_\mathbf{x} \log p_t(\mathbf{x})\bigr]\,dt
    + g(t)\,d\bar{\mathbf{W}},
\end{equation}
where $\nabla_\mathbf{x} \log p_t(\mathbf{x})$ is the \emph{score function}
of the marginal density at time $t$.  A neural network $s_\theta(\mathbf{x},t)$
is trained to approximate this score via denoising score matching:
\begin{equation}
  \mathcal{L}_\text{DSM} =
    \E_{t,\mathbf{x}_0, \boldsymbol\varepsilon}\!\left[
      \lambda(t)\,\bigl\|s_\theta(\mathbf{x}_t, t)
        + \tfrac{\boldsymbol\varepsilon}{\sigma_t}\bigr\|^2\right],
\end{equation}
where $\mathbf{x}_t = \sqrt{\bar\alpha_t}\,\mathbf{x}_0 + \sqrt{1-\bar\alpha_t}\,\boldsymbol\varepsilon$
and $\boldsymbol\varepsilon \sim \mathcal{N}(0,I)$.  This loss has a simple
probabilistic interpretation: the network should predict the direction of
increasing data density from any noisy observation.

\paragraph{From score matching to bridge matching.}
Flow Matching~\cite{lipman2022flow} and stochastic interpolants~\cite{albergo2023stochastic}
generalise this by directly specifying an interpolation path
$\mathbf{x}_t = \alpha(t)\,\mathbf{x}_1 + (1-\alpha(t))\,\mathbf{x}_0 + \sigma(t)\,\boldsymbol\varepsilon$
between a source $\mathbf{x}_0 \sim \mathcal{N}(0,I)$ and a target
$\mathbf{x}_1 \sim p_\text{data}$, and training a velocity network to match
the instantaneous derivative $\dot{\mathbf{x}}_t$.  The choice of
interpolation schedules $\alpha(t)$ and $\sigma(t)$ controls the geometry
of the bridge; cosine and sinusoidal schedules have been shown empirically to
produce smoother training signals and better-conditioned velocity
fields~\cite{lipman2022flow}.

Crucially, \emph{none of these formulations assign differential importance to
points along the bridge}.  A bridge sample at $t=0.1$ receives the same weight
in the regression loss as one at $t=0.9$, and a sample from a
high-density region of the data manifold receives the same weight as one from
a degenerate or ambiguous transition.  This is the gap AGM addresses.

\section{Methodology}
\label{sec:method}

\subsection{Stochastic Bridge Construction}

Let $\mathbf{x}_0 \sim \mathcal{N}(0,I)$ denote a noise sample and
$\mathbf{x}_1 \sim p_\text{data}$ a real image.  We construct a stochastic
bridge interpolating between them as
\begin{equation}
  \mathbf{x}_t = \alpha(t)\,\mathbf{x}_1
    + \bigl(1-\alpha(t)\bigr)\,\mathbf{x}_0
    + \sigma(t)\,\boldsymbol\varepsilon,
  \qquad \boldsymbol\varepsilon \sim \mathcal{N}(0,I),
  \label{eq:bridge}
\end{equation}
where $t \sim \mathcal{U}(0,1)$ is sampled uniformly at each training step.
We use \emph{cosine interpolation} and a \emph{sinusoidal bridge width}:
\begin{align}
  \alpha(t) &= \sin^2\!\left(\tfrac{\pi t}{2}\right),
  \label{eq:alpha} \\
  \sigma(t) &= \sigma_{\max} \cdot \sin(\pi t),
  \label{eq:sigma}
\end{align}
with $\sigma_{\max} = 0.01$ for the publication run.
The cosine schedule ensures $\alpha(0)=0$ and $\alpha(1)=1$ with a
smooth sigmoid-like transition that avoids sharp regime boundaries.
The sinusoidal bridge width $\sigma(t)$ vanishes at both endpoints---so the
boundary conditions are preserved exactly---and reaches its maximum at
$t=0.5$, adding the most stochasticity at midpoint where the interpolant is
maximally uncertain.

The \emph{target velocity} for the bridge, obtained by differentiating
Eq.~\eqref{eq:bridge} with respect to $t$, is
\begin{equation}
  \dot{\mathbf{x}}_t
    = \alpha'(t)\,(\mathbf{x}_1 - \mathbf{x}_0)
    + \sigma'(t)\,\boldsymbol\varepsilon,
  \label{eq:target_vel}
\end{equation}
where $\alpha'(t) = \tfrac{\pi}{2}\sin(\pi t)$ and
$\sigma'(t) = \sigma_{\max}\pi\cos(\pi t)$.
A drift network $f_\theta(\mathbf{x}_t, t)$ is trained to regress this
target velocity, so that at inference one can integrate from
$\mathbf{x}_0 \sim \mathcal{N}(0,I)$ to $\mathbf{x}_1 \approx p_\text{data}$
by numerically solving $d\mathbf{x}/dt = f_\theta(\mathbf{x}_t, t)$.

\subsection{The Potential Network $V_\phi$}

The central contribution of AGM is the introduction of a learned scalar
potential $V_\phi : \R^{H\times W \times C} \to \R$ that operates on bridge
samples $\mathbf{x}_t$ and outputs a signed scalar score reflecting how
structurally informative that sample is.

\paragraph{Objective.}
$V_\phi$ is trained with a \emph{hinge-margin contrastive loss} that pushes
bridge samples $\mathbf{x}_t$ (drawn from real data bridges) to have high
potential, and pushes pure Gaussian noise $\mathbf{z} \sim \mathcal{N}(0,I)$
to have low potential, with a margin of $m$ between them:
\begin{equation}
  \mathcal{L}_V =
    \E_{\mathbf{x}_t}\!\left[\operatorname{ReLU}\!\left(V_\phi(\mathbf{x}_t) + m\right)\right]
    + \E_{\mathbf{z}}\!\left[\operatorname{ReLU}\!\left(m - V_\phi(\mathbf{z})\right)\right]
    + \gamma_V \,\E_{\mathbf{x}_t}\!\left[V_\phi(\mathbf{x}_t)^2\right],
  \label{eq:loss_v}
\end{equation}
where $m=1.0$ is the margin and $\gamma_V = 10^{-3}$ is an $L_2$ regulariser
preventing potential explosion.  The first term penalises bridge samples that
score too low (potential $< -m$); the second penalises noise samples that
score too high ($> m$); the third regularises the scale of $V_\phi$'s outputs.

Intuitively, $V_\phi$ learns to answer the question: ``does this point in
the bridge landscape look more like real-data structure or like noise?''
Samples near the data manifold attract high positive potential;
off-manifold transitions attract lower or negative potential.

\subsection{Importance Weighting with Stop-Gradient}

Given $V_\phi$'s scores on a minibatch of bridge samples, we convert them
into non-negative importance weights.  Concretely, let
$\sigma(\cdot)$ denote the sigmoid function.
The weight for sample $\mathbf{x}_t^{(i)}$ in a batch is
\begin{equation}
  w^{(i)} = 1 + \lambda_V
    \left(
      \frac{\sigma\bigl(\sg[V_\phi(\mathbf{x}_t^{(i)})]\bigr)}
           {\E_\text{batch}\bigl[\sigma\bigl(\sg[V_\phi]\bigr)\bigr]}
      - 1
    \right),
  \label{eq:weight}
\end{equation}
where $\sg[\cdot]$ is the \emph{stop-gradient} operator (implemented as
\texttt{.detach()} in PyTorch), $\lambda_V = 0.1$ controls the strength of
the re-weighting, and the batch-level normalisation ensures the weights have
unit mean, so the expected gradient magnitude is preserved.  The weight
$w^{(i)}$ is clamped to $[0.1, 10.0]$ to prevent numerical instability.

\subsection{Why the Stop-Gradient is Essential}
\label{sec:sg}

Without $\sg[\cdot]$, gradients from $\mathcal{L}_f$ flow back through the
importance weights into $V_\phi$, and this creates a specific, analysable
failure mode rather than merely noisy training.  To see why, expand the
gradient of the drift loss with respect to $V_\phi$.  Dropping the batch
normalisation for clarity, the weight $w^{(i)}$ depends on $V_\phi$ via
$w^{(i)} \propto \sigma(V_\phi(\mathbf{x}_t^{(i)}))$, so
\begin{equation}
  \frac{\partial \mathcal{L}_f}{\partial V_\phi}
    = \E\!\left[
        \bigl\|f_\theta(\mathbf{x}_t,t) - \dot{\mathbf{x}}_t\bigr\|^2
        \cdot \lambda_V\,\sigma'(V_\phi(\mathbf{x}_t))\,
        \nabla_{V_\phi}
      \right].
  \label{eq:sg_grad}
\end{equation}
This gradient incentivises $V_\phi$ to \emph{minimise the drift loss
directly}: it can reduce $\mathcal{L}_f$ most efficiently by assigning
low potential---and therefore low weight---to whichever bridge samples are
currently hard for $f_\theta$, irrespective of whether those samples are
structurally uninformative.  The result is a degenerate equilibrium in
which $V_\phi$ functions as a gating network that suppresses difficult
examples rather than a potential that scores structural salience.  Because
$f_\theta$ simultaneously adapts to this re-weighting, the two networks
enter an adversarial spiral: $V_\phi$ learns to down-weight the transitions
where $f_\theta$ is weakest, $f_\theta$ never improves on those transitions,
and the loss diverges.  In practice we observe training collapse within
20{,}000 steps when \texttt{.detach()} is omitted.

The stop-gradient operator $\sg[\cdot]$ severs Eq.~\eqref{eq:sg_grad}
entirely.  $V_\phi$ receives gradient signal only from $\mathcal{L}_V$,
which drives it to score bridge samples versus noise.  Its scalar output
reaches $f_\theta$ solely as a detached numerical weight---an asymmetric
information channel that preserves the guiding signal whilst eliminating
the adversarial shortcut.

\subsection{Drift Loss and Training Procedure}

The drift network is trained on an importance-weighted bridge-matching
objective:
\begin{equation}
  \mathcal{L}_f =
    \E_{t, \mathbf{x}_0, \mathbf{x}_1, \boldsymbol\varepsilon}\!\left[
      w \cdot \bigl\|f_\theta(\mathbf{x}_t, t) - \dot{\mathbf{x}}_t\bigr\|^2
    \right],
  \label{eq:loss_f}
\end{equation}
where $\dot{\mathbf{x}}_t$ is the target velocity from Eq.~\eqref{eq:target_vel}
and $w$ is the stop-gradient importance weight from Eq.~\eqref{eq:weight}.

The training order within each step matters: $V_\phi$ must be updated
\emph{before} computing the importance weights for the drift update,
so that the weights reflect the most current potential estimate.

\begin{algorithm}[t]
  \caption{AGM Training Step}
  \label{alg:train}
  \begin{algorithmic}[1]
    \Require Batch of real images $\{\mathbf{x}_1^{(i)}\}$; networks $V_\phi$, $f_\theta$;
      EMA model $\bar{f}_\theta$; hyperparameters $m, \gamma_V, \lambda_V, \sigma_{\max}$
    \State Sample $\mathbf{x}_0^{(i)} \sim \mathcal{N}(0,I)$,\; $t^{(i)} \sim \mathcal{U}(0,1)$,\;
      $\boldsymbol\varepsilon^{(i)} \sim \mathcal{N}(0,I)$
    \State Construct bridge: $\mathbf{x}_{t}^{(i)} \leftarrow \alpha(t^{(i)})\mathbf{x}_1^{(i)}
      + (1-\alpha(t^{(i)}))\mathbf{x}_0^{(i)} + \sigma(t^{(i)})\boldsymbol\varepsilon^{(i)}$
    \State \textbf{(1) Update $V_\phi$:}
      Sample noise $\mathbf{z}^{(i)} \sim \mathcal{N}(0,I)$;
      compute $\mathcal{L}_V$ via Eq.~\eqref{eq:loss_v}; step optimiser
    \State \textbf{(2) Compute weights:}
      $w^{(i)} \leftarrow$ Eq.~\eqref{eq:weight} using $\sg[V_\phi(\mathbf{x}_t^{(i)})]$
    \State \textbf{(3) Update $f_\theta$:}
      Compute target velocity $\dot{\mathbf{x}}_t^{(i)}$ via Eq.~\eqref{eq:target_vel};
      compute $\mathcal{L}_f$ via Eq.~\eqref{eq:loss_f}; step optimiser
    \State \textbf{(4) EMA step:}
      $\bar{f}_\theta \leftarrow \tau\,\bar{f}_\theta + (1-\tau)\,f_\theta$,\;
      $\tau = 0.9999$
  \end{algorithmic}
\end{algorithm}

\begin{algorithm}[t]
  \caption{AGM Sampling (EMA drift only; $V_\phi$ discarded)}
  \label{alg:sample}
  \begin{algorithmic}[1]
    \Require EMA model $\bar{f}_\theta$; NFE $N$; inference noise $\sigma_\text{SDE}$
    \State $\mathbf{x}_0 \sim \mathcal{N}(0,I)$;\quad $\Delta t \leftarrow 1/N$
    \For{$n = 0, 1, \ldots, N-1$}
      \State $t \leftarrow n\,\Delta t$
      \State $\mathbf{x}_{t+\Delta t} \leftarrow \mathbf{x}_t
        + \bar{f}_\theta(\mathbf{x}_t, t)\,\Delta t
        + \sigma_\text{SDE}\,\sqrt{\Delta t}\,\boldsymbol\xi$,\quad
        $\boldsymbol\xi \sim \mathcal{N}(0,I)$
        \quad\Comment{Euler--Maruyama}
    \EndFor
    \State \Return $\mathbf{x}_1$
  \end{algorithmic}
\end{algorithm}

\subsection{Network Architectures}

\paragraph{Drift U-Net $f_\theta$.}
The drift network is a U-Net with three resolution scales: $64 \to 32 \to 16
\to 8$ pixels.  Each scale contains two residual blocks (ResBlocks) with group
normalisation and SiLU activations.  Self-attention is applied at the two
innermost scales ($16\times16$ and $8\times8$) and at the bottleneck.  The
base channel width is $C=128$, giving approximately 50M parameters.  Time
conditioning uses sinusoidal positional embeddings projected through a
two-layer Multi-Layer Perceptron (MLP); the resulting time vector is injected into every ResBlock via
FiLM (feature-wise linear modulation), providing scale and shift parameters.
The output convolution is zero-initialised to ensure the velocity field starts
near zero, stabilising early training.  An exponential moving average (EMA) with decay $\tau = 0.9999$ is
maintained throughout training and used exclusively at inference.

\paragraph{Potential network $V_\phi$.}
$V_\phi$ is intentionally lightweight: three strided convolutions (stride 2,
kernel 3) with GroupNorm and Gaussian Error Linear Unit (GELU) activations, followed by adaptive average
pooling to a $4\times4$ spatial feature map, flattened and projected to a
scalar output.  The network has approximately 0.7M parameters---about 1.4\%
of $f_\theta$---so its computational overhead during training is negligible.
$V_\phi$ receives no time conditioning; it operates on the bridge sample
$\mathbf{x}_t$ directly.  The absence of time input is a deliberate design
choice: we want $V_\phi$ to learn a time-marginal notion of structural
informativeness, not a time-specific one.

\paragraph{Why $V_\phi$ is discarded at inference.}
$V_\phi$ serves a purely training-time role: its output shapes the gradient
signal that $f_\theta$ receives.  Once training is complete, the EMA drift
$\bar{f}_\theta$ has already absorbed $V_\phi$'s guidance into its weights.
Running $V_\phi$ at inference would provide no additional information
(it is not part of the generative SDE) and would add latency with no benefit.
Its discard is therefore not a design compromise but an inherent property of
the framework: AGM's inference graph is exactly that of a standard bridge-matching
model.

\subsection{Architecture Diagrams}

\begin{figure}[htbp]
\centering
\resizebox{0.97\linewidth}{!}{%
\begin{tikzpicture}[
  >=Stealth,
  every node/.style={font=\normalsize},
  dn/.style={ellipse, draw=gray!55, fill=agmgraybg, line width=0.9pt,
             minimum width=3.4cm, minimum height=0.95cm, align=center},
  loopbox/.style={rectangle, rounded corners=8pt,
                  draw=agmblue, fill=agmbluebg, line width=1.2pt,
                  align=center, inner sep=10pt},
  dead/.style={rectangle, rounded corners=6pt,
               draw=gray!55, fill=gray!22, line width=0.8pt,
               align=center, inner sep=7pt, text=gray!70},
  ma/.style={-Stealth, line width=1.0pt, color=gray!65},
  meb/.style={rectangle, rounded corners=1.5pt,
              draw=agmblue!60, fill=agmblue!18, line width=0.5pt,
              minimum height=0.3cm, inner sep=2pt,
              font=\scriptsize\bfseries\color{agmblue!90}},
  mdb/.style={rectangle, rounded corners=1.5pt,
              draw=agmblue!50, fill=agmblue!10, line width=0.5pt,
              minimum height=0.3cm, inner sep=2pt,
              font=\scriptsize\bfseries\color{agmblue!80}},
  mbot/.style={rectangle, rounded corners=1.5pt,
               draw=agmblue!80, fill=agmblue!32, line width=0.5pt,
               minimum height=0.3cm, inner sep=2pt,
               font=\scriptsize\bfseries\color{agmblue!95}},
  sk/.style={-Stealth, line width=0.35pt, draw=gray!35, dashed},
]

\node[dn] (noise) at (0,0)
  {$\mathbf{x}_0 \sim \mathcal{N}(\mathbf{0},\mathbf{I})$\\[-3pt]
   \footnotesize $t = 0$ \;(pure noise)};

\node[loopbox] (loop) at (7.5,0) {%
  \begin{tikzpicture}[inner sep=0, baseline=(base.base)]
    \node[meb, minimum width=1.5cm] (me1) at (-1.25, 0.72)  {$64{\times}64$};
    \node[meb, minimum width=1.25cm](me2) at (-1.25, 0.22)  {$32{\times}32$};
    \node[meb, minimum width=1.0cm] (me3) at (-1.25,-0.28)  {$16{\times}16$};
    \node[mbot,minimum width=0.75cm](mbo) at (0.0,  -0.78)  {$8{\times}8$};
    \node[mdb, minimum width=1.0cm] (md3) at ( 1.25,-0.28)  {$16{\times}16$};
    \node[mdb, minimum width=1.25cm](md2) at ( 1.25, 0.22)  {$32{\times}32$};
    \node[mdb, minimum width=1.5cm] (md1) at ( 1.25, 0.72)  {$64{\times}64$};
    \draw[-Stealth,agmblue!50,line width=0.4pt] (me1)--(me2);
    \draw[-Stealth,agmblue!50,line width=0.4pt] (me2)--(me3);
    \draw[-Stealth,agmblue!50,line width=0.4pt] (me3.south) |- (mbo.west);
    \draw[-Stealth,agmblue!50,line width=0.4pt] (mbo.east) -| (md3.south);
    \draw[-Stealth,agmblue!50,line width=0.4pt] (md3)--(md2);
    \draw[-Stealth,agmblue!50,line width=0.4pt] (md2)--(md1);
    \draw[sk] (me1.east) -- (md1.west);
    \draw[sk] (me2.east) -- (md2.west);
    \draw[sk] (me3.east) -- (md3.west);
    \node[font=\footnotesize\bfseries\color{agmblue!80}] at (0,1.2)
      {$\bar{f}_\theta(\mathbf{x}_t,t)$};
    \coordinate (base) at (0,0);
  \end{tikzpicture}\\[5pt]
  {\normalsize\textbf{Euler--Maruyama}\;($N$ steps,\;$t:0\!\to\!1$)}\\[3pt]
  {\small$\mathbf{x}_{t+\Delta t}
     = \mathbf{x}_t
     + \bar{f}_\theta(\mathbf{x}_t,t)\,\Delta t
     + \sigma_\text{SDE}\sqrt{\Delta t}\;\boldsymbol{\xi}$}
};

\node[dn] (gen) at (15.5,0)
  {$\mathbf{x}_1 \approx p_{\mathrm{data}}$\\[-3pt]
   \footnotesize $t = 1$ \;(generated image)};

\node[dead] (dead) at (7.5,-4.0)
  {$V_\phi$\quad\textbf{discarded at inference}\\[-2pt]
   \footnotesize no computational overhead whatsoever};

\draw[ma] (noise.east) -- (loop.west)
  node[midway, above, font=\small, color=gray!60]{$t=0$};
\draw[ma] (loop.east) -- (gen.west)
  node[midway, above, font=\small, color=gray!60]{$t=1$};

\draw[-Stealth, line width=1.0pt, color=agmblue!65]
  ($(loop.north)+(-1.2,0)$) arc[start angle=200, end angle=-20, radius=0.95cm]
  node[pos=0.5, above=6pt, font=\small, color=agmblue!70]{step $\Delta t$};

\draw[agmred!45, line width=1.3pt] (dead.north west) -- (dead.south east);
\draw[agmred!45, line width=1.3pt] (dead.north east) -- (dead.south west);

\draw[gray!30, dashed, line width=0.6pt] (dead.north) -- (loop.south);

\end{tikzpicture}%
}%
\caption{%
  \textbf{AGM sampling pipeline.}
  Starting from isotropic Gaussian noise at $t=0$, the EMA drift
  $\bar{f}_\theta$ (shown as a mini U-Net schematic) is integrated for $N$
  Euler--Maruyama steps.  The potential network $V_\phi$ plays no role
  whatsoever at inference and is entirely discarded.%
}
\label{fig:sample_diag}
\end{figure}
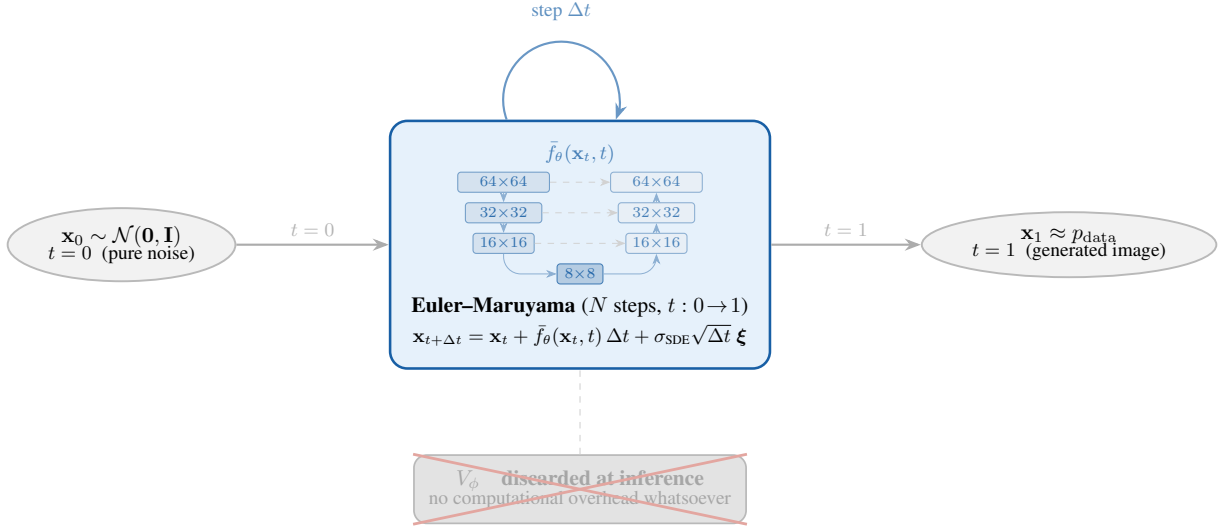

Figures~\ref{fig:train_diag} and~\ref{fig:sample_diag} illustrate the
training and sampling pipelines respectively.

\begin{figure}[htbp]
\centering
\resizebox{\linewidth}{!}{%
\begin{tikzpicture}[
  >=Stealth,
  every node/.style={font=\normalsize},
  dn/.style={ellipse, draw=gray!55, fill=agmgraybg, line width=0.9pt,
             minimum width=3.2cm, minimum height=0.9cm, align=center},
  gb/.style={rectangle, rounded corners=5pt, draw=gray!50, fill=white,
             line width=0.8pt, align=center, inner sep=7pt},
  eb/.style={rectangle, rounded corners=2pt,
             draw=agmblue!65, fill=agmblue!18, line width=0.7pt,
             minimum height=0.44cm, align=center, inner sep=3pt,
             font=\footnotesize\bfseries\color{agmblue!90}},
  db/.style={rectangle, rounded corners=2pt,
             draw=agmblue!55, fill=agmblue!10, line width=0.7pt,
             minimum height=0.44cm, align=center, inner sep=3pt,
             font=\footnotesize\bfseries\color{agmblue!90}},
  bot/.style={rectangle, rounded corners=2pt,
              draw=agmblue!80, fill=agmblue!35, line width=0.8pt,
              minimum height=0.44cm, align=center, inner sep=3pt,
              font=\footnotesize\bfseries\color{agmblue!95}},
  lb/.style={rectangle, rounded corners=4pt,
             draw=agmgreen!70, fill=agmgreenbg, line width=0.8pt,
             align=center, inner sep=5pt,
             font=\normalsize\color{agmgreen!90}},
  emab/.style={rectangle, rounded corners=4pt,
               draw=agmblue!70, fill=agmbluebg, line width=0.8pt,
               align=center, inner sep=5pt,
               font=\normalsize\color{agmblue}},
  wb/.style={rectangle, rounded corners=4pt,
             draw=agmred!55, fill=agmred!7, line width=0.8pt,
             align=center, inner sep=5pt},
  ma/.style={-Stealth, line width=1.0pt, color=gray!65},
  sga/.style={-Stealth, line width=1.1pt, draw=agmred, dashed},
  sk/.style={-Stealth, line width=0.45pt, draw=gray!40, dashed},
  an/.style={circle, draw=agmamber!80, fill=agmamber!22,
             line width=0.55pt, minimum size=8pt, inner sep=0},
  vb/.style={rectangle, rounded corners=2pt,
             draw=agmamber!70, fill=agmamber!15, line width=0.7pt,
             minimum height=0.5cm, align=center, inner sep=4pt,
             font=\footnotesize\bfseries\color{agmamber!90}},
]

\node[dn] (x0) at (-5.5, 0) {$\mathbf{x}_0 \sim \mathcal{N}(\mathbf{0},\mathbf{I})$};
\node[dn] (x1) at ( 5.5, 0) {$\mathbf{x}_1 \sim p_{\mathrm{data}}$};

\node[gb, text width=10.5cm] (bridge) at (0,-1.75)
  {$\mathbf{x}_t \;=\; \alpha(t)\,\mathbf{x}_1
      \,+\,\bigl(1-\alpha(t)\bigr)\,\mathbf{x}_0
      \,+\,\sigma(t)\,\boldsymbol{\varepsilon},
   \quad t\sim\mathcal{U}(0,1)$};
\draw[ma] (x0.east) -| ($(bridge.north west)!0.25!(bridge.north east)$);
\draw[ma] (x1.west) -| ($(bridge.north east)!0.25!(bridge.north west)$);

\draw[rounded corners=10pt, draw=agmamber, fill=agmamberbg, line width=1.3pt]
  (-8.6,-3.0) rectangle (-0.4,-9.4);
\draw[rounded corners=10pt, draw=agmblue,  fill=agmbluebg,  line width=1.3pt]
  ( 0.4,-3.0) rectangle ( 9.1,-9.4);

\node[font=\large\bfseries\color{agmamber}]           at (-4.5,-3.45) {$V_\phi$ \;---\; PotentialNet};
\node[font=\small\itshape\color{agmamber!75}]          at (-4.5,-3.95) {(stride-2 CNN,\;$\sim$0.7M params;\;updated \textbf{first})};
\node[font=\large\bfseries\color{agmblue}]             at ( 4.75,-3.45) {$f_\theta$ \;---\; DriftUNet};
\node[font=\small\itshape\color{agmblue!80}]            at ( 4.75,-3.95) {(U-Net encoder--decoder,\;$\sim$50M params;\;updated \textbf{second})};

\draw[line width=1.0pt, color=gray!65] (bridge.south) -- (0,-2.65);
\draw[ma] (0,-2.65) -| (-4.5,-3.0);
\draw[ma] (0,-2.65) -| ( 4.75,-3.0);

\begin{scope}[shift={(-4.5,-6.0)}]
  \node[vb, minimum width=3.6cm] (vc1) at (0, 0.8)
    {Conv\,+\,GN\;(stride-2)};
  \node[vb, minimum width=3.6cm] (vc2) at (0, 0.0)
    {Conv\,+\,GN\;(stride-2)};
  \node[vb, minimum width=3.6cm] (vc3) at (0,-0.8)
    {Conv\,+\,GN\;(stride-2)};
  \node[vb, minimum width=3.8cm] (vpl) at (0,-1.7)
    {AvgPool\;$4{\times}4$ + Linear};
  \draw[ma] (vc1)--(vc2);
  \draw[ma] (vc2)--(vc3);
  \draw[ma] (vc3)--(vpl);
  \node[font=\footnotesize\bfseries\color{agmamber}] at (0,-2.45)
    {$\to V_\phi(\mathbf{x}_t)\in\mathbb{R}$};
  \coordinate (vphi_out) at (0,-1.7);
\end{scope}

\begin{scope}[shift={(4.75,-7.0)}]

  \node[eb, minimum width=2.2cm] (e1) at (-1.5, 1.7)  {$64{\times}64$,\,$C$};
  \node[eb, minimum width=1.85cm](e2) at (-1.5, 0.75) {$32{\times}32$,\,$2C$};
  \node[eb, minimum width=1.5cm] (e3) at (-1.5,-0.2)  {$16{\times}16$,\,$4C$\;{\footnotesize$\star$}};

  \node[bot, minimum width=1.15cm] (b) at (0,-1.25) {$8{\times}8$,\,$8C$\;{\footnotesize$\star$}};

  \node[db, minimum width=1.5cm] (d3) at (1.5,-0.2)  {$16{\times}16$,\,$4C$\;{\footnotesize$\star$}};
  \node[db, minimum width=1.85cm](d2) at (1.5, 0.75) {$32{\times}32$,\,$2C$};
  \node[db, minimum width=2.2cm] (d1) at (1.5, 1.7)  {$64{\times}64$,\,$C$};

  \draw[ma] (e1.south) -- (e2.north);
  \draw[ma] (e2.south) -- (e3.north);
  \draw[ma] (e3.south) |- (b.west);

  \draw[ma] (b.east) -| (d3.south);
  \draw[ma] (d3.north) -- (d2.south);
  \draw[ma] (d2.north) -- (d1.south);

  \draw[sk] (e1.east) -- (d1.west) node[midway, above, font=\tiny, color=gray!50]{skip};
  \draw[sk] (e2.east) -- (d2.west) node[midway, above, font=\tiny, color=gray!50]{skip};
  \draw[sk] (e3.east) -- (d3.west) node[midway, above, font=\tiny, color=gray!50]{skip};

  \node[font=\small\color{agmblue!65}, align=center] (te) at (-3.2, 0.75)
    {$t$\,emb.\\FiLM};
  \draw[-Stealth, agmblue!40, line width=0.55pt] (te.east) -- (e2.west);

  \node[font=\scriptsize\color{gray!50}] at (2.6,-0.8)
    {{\footnotesize$\star$} self-attn};

  \coordinate (ftheta_out_s) at (0,-1.9);
  \coordinate (ftheta_out_e) at (3.7, 0.75);

\end{scope}

\node[wb, text width=8.0cm] (wbox) at (-0.5,-11.0)
  {\normalsize
   $w^{(i)} = 1 + \lambda_V\!\left(
      \dfrac{\sigma\!\bigl(\sg[V_\phi(\mathbf{x}_t^{(i)})]\bigr)}
            {\mathbb{E}_\text{batch}\!\bigl[\sigma(\sg[V_\phi])\bigr]}
      - 1\right)$};

\draw[sga]
  (-4.5,-9.4) to[out=270, in=180, looseness=1.4]
  node[pos=0.28, xshift=-0.9cm, font=\small, color=agmred]{$\sg[\cdot]$}
  (wbox.west);

\draw[sga]
  (wbox.east) -| (6.5,-9.4)
  node[pos=0.93, right=4pt, font=\small, color=agmred]{weighted $w$};

\node[lb] (lv) at (-4.5,-12.6) {$\mathcal{L}_V$\;(Eq.~\ref{eq:loss_v})};
\node[lb] (lf) at ( 4.75,-12.6){$\mathcal{L}_f$\;(Eq.~\ref{eq:loss_f})};
\node[emab] (ema) at (12.2,-6.35)
  {EMA $\bar{f}_\theta$\\[3pt]\small inference\\[-1pt]\small model};

\draw[ma] (wbox.south) |- (lv.east);
\draw[ma] (4.75,-9.4) -- (lf.north);

\draw[-Stealth, line width=0.95pt, color=agmblue!70]
  (9.1,-6.35) -- (ema.west)
  node[midway, above, font=\small, color=agmblue!75]{$\tau{=}0.9999$};

\end{tikzpicture}%
}%
\caption{%
  \textbf{AGM training pipeline.}
  \textbf{Left (amber):} $V_\phi$ (PotentialNet) shown as a
  \emph{block diagram} (three stride-2 Conv+GN layers followed by adaptive
  AvgPool and a linear projection to a scalar output).
  \textbf{Right (blue):} $f_\theta$ (DriftUNet) shown as a
  standard \emph{U-Net encoder--bottleneck--decoder block diagram} with
  skip connections (dashed) and self-attention at inner scales ($\star$).
  The stop-gradient barrier (red dashed, $\sg[\cdot]$) ensures
  $V_\phi$'s scalar output forms importance weights $w$ without
  back-propagating gradients into $\mathcal{L}_f$.
  Only the EMA snapshot $\bar{f}_\theta$ is retained for inference.%
}
\label{fig:train_diag}
\end{figure}
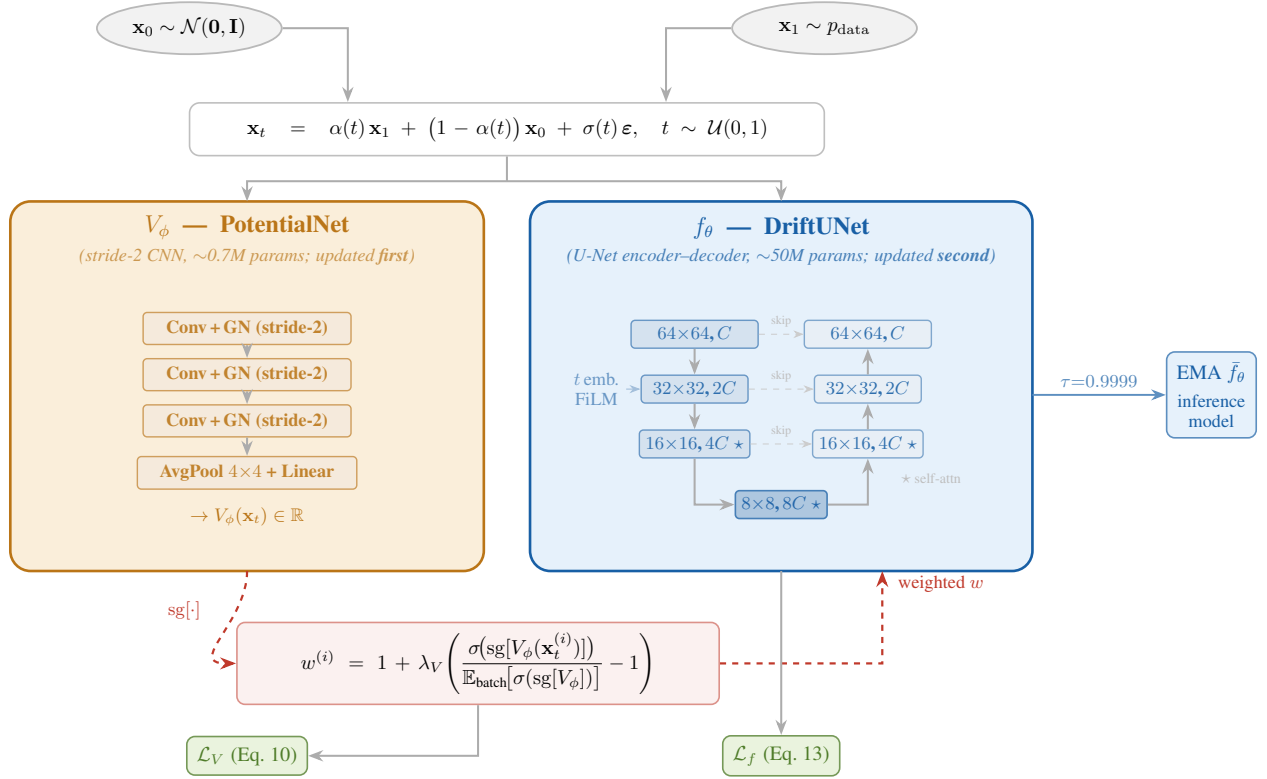

\section{Experimental Setup}
\label{sec:setup}

\paragraph{Dataset.}
We train on CelebA-HQ~\cite{karras2018progressive} at $64{\times}64$ resolution,
comprising approximately 30{,}000 face images streamed from the
\texttt{huggan/celeba-hq} HuggingFace dataset.

\paragraph{Hyperparameters.}
Table~\ref{tab:hparams} lists the full hyperparameter configuration used
for the publication run.

\begin{table}[ht]
  \caption{Hyperparameter configuration for the 500k-step publication run.}
  \label{tab:hparams}
  \centering
  \small
  \begin{tabular}{ll}
    \toprule
    Hyperparameter & Value \\
    \midrule
    Training steps          & 500{,}000 \\
    Batch size              & 128 \\
    Image resolution        & $64{\times}64$ \\
    Base channels $C$       & 128 \\
    Time embedding dim      & 256 \\
    Learning rate ($f_\theta$) & $2{\times}10^{-4}$ \\
    Learning rate ($V_\phi$)   & $1{\times}10^{-4}$ \\
    EMA decay $\tau$        & 0.9999 \\
    Inference noise $\sigma_\text{SDE}$ & 0.01 \\
    Bridge margin $m$       & 1.0 \\
    $V_\phi$ regulariser $\gamma_V$ & $10^{-3}$ \\
    Importance weight $\lambda_V$ & 0.1 \\
    Weight clamp range      & $[0.1,\, 10.0]$ \\
    NFE at evaluation       & 500 \\
    FID sample count        & 10{,}000 \\
    \bottomrule
  \end{tabular}
\end{table}

\paragraph{Compute.}
The full 500{,}000-step training run required approximately 36 hours on a
single NVIDIA RTX 4090 (24\,GB VRAM) with mixed-precision training
(\texttt{torch.amp.autocast}).  We emphasise that this is a
\emph{single-consumer-GPU} result.
Published models achieving single-digit FID---DDPM reports
FID~$\approx3$~\cite{ho2020ddpm}, Score SDE~$\approx2$~\cite{song2021score}---
used 8 to 512 TPUs or A100s, 500k to 1.3M training steps,
and training sets of 162k or more images.
A direct FID comparison would be misleading; the relevant comparison is the
\emph{controlled ablation} reported in Section~\ref{sec:results}.

\section{Results}
\label{sec:results}

\subsection{Ablation: Effect of $V_\phi$}

Table~\ref{tab:ablation} reports the core controlled experiment: a drift-only
baseline ($f_\theta$ with $\lambda_V = 0$, all other settings identical)
versus the full AGM ($f_\theta + V_\phi$), both trained for 500{,}000 steps
under the same configuration on a single NVIDIA RTX 4090.  The absolute Fréchet Inception Distance (FID)
numbers are not the point here---they reflect the modest compute budget and
dataset size, and should not be compared against published results that used
8 to 512 TPUs and tenfold more training data.  What matters is the
\emph{relative improvement} that is attributable exclusively to the
$V_\phi$ mechanism, since every other factor---hardware, data, batch size,
learning rates, architecture, training duration---is held fixed.  Under
those controlled conditions, introducing the stop-gradient potential reduces
FID by 10.7\%, improves precision by 2.8\%, and improves recall by 2.1\%.
The gain is consistent across all three metrics, which together span both
fidelity and coverage; the potential does not buy one at the expense of the
other.

We note that the three AGM-specific hyperparameters---the importance weight
strength $\lambda_V$, the hinge margin $m$, and the regulariser $\gamma_V$---were
set by principled criteria rather than swept against FID.  $\lambda_V = 0.1$
was chosen to keep importance weights in a unit-mean regime via the
batch-level normalisation in Eq.~\eqref{eq:weight}; $m = 1.0$ follows
standard practice for hinge-margin contrastive objectives; $\gamma_V =
10^{-3}$ is a light $L_2$ term that prevents potential explosion without
substantially constraining the landscape.  We do not claim this configuration
is globally optimal.  The claim is narrower and internally valid: under
identical conditions, the $V_\phi$ mechanism---not a hyperparameter
advantage---is responsible for the observed improvement.

\begin{table}[ht]
  \caption{%
    Ablation of $V_\phi$ on CelebA-HQ $64{\times}64$.
    Both variants trained for 500{,}000 steps on the same hardware.
    FID computed on 10{,}000 samples at NFE$=500$.%
  }
  \label{tab:ablation}
  \centering
  \begin{tabular}{lccc}
    \toprule
    Model & FID\,$\downarrow$ & Precision\,$\uparrow$ & Recall\,$\uparrow$ \\
    \midrule
    $f_\theta$ only (no $V_\phi$) & 28.53 & 0.705 & 0.932 \\
    \textbf{Full AGM} ($f_\theta + V_\phi$) & \textbf{25.47} & \textbf{0.725} & \textbf{0.952} \\
    \midrule
    $\Delta$ (absolute) & $-3.06$ & $+0.020$ & $+0.020$ \\
    $\Delta$ (relative) & $-10.7\%$ & $+2.8\%$ & $+2.1\%$ \\
    \bottomrule
  \end{tabular}
\end{table}

\subsection{Generation Quality Metrics}

Table~\ref{tab:pr} summarises the precision and recall of the full AGM.
Recall of 0.952 indicates that the generated distribution covers nearly all
of the real data manifold---there is minimal mode collapse.
Precision of 0.725 indicates that approximately 72.5\% of generated samples
fall within the support of the real distribution, with room for further
improvement through longer training or architectural scaling.

\begin{table}[ht]
  \caption{%
    Precision and recall of the full AGM on CelebA-HQ $64{\times}64$
    (NFE$=500$, 10{,}000 generated samples).%
  }
  \label{tab:pr}
  \centering
  \begin{tabular}{lcc}
    \toprule
    Model & Precision\,$\uparrow$ & Recall\,$\uparrow$ \\
    \midrule
    Full AGM & 0.725 & 0.952 \\
    \bottomrule
    \multicolumn{3}{l}{\small\textit{Recall $\geq 0.85$ confirms strong mode coverage with minimal collapse.}} \\
    \multicolumn{3}{l}{\small\textit{Precision is expected to improve with additional compute.}} \\
  \end{tabular}
\end{table}

\subsection{Visual Quality}

\begin{figure}[htbp]
  \centering
  \includegraphics[width=0.82\linewidth]{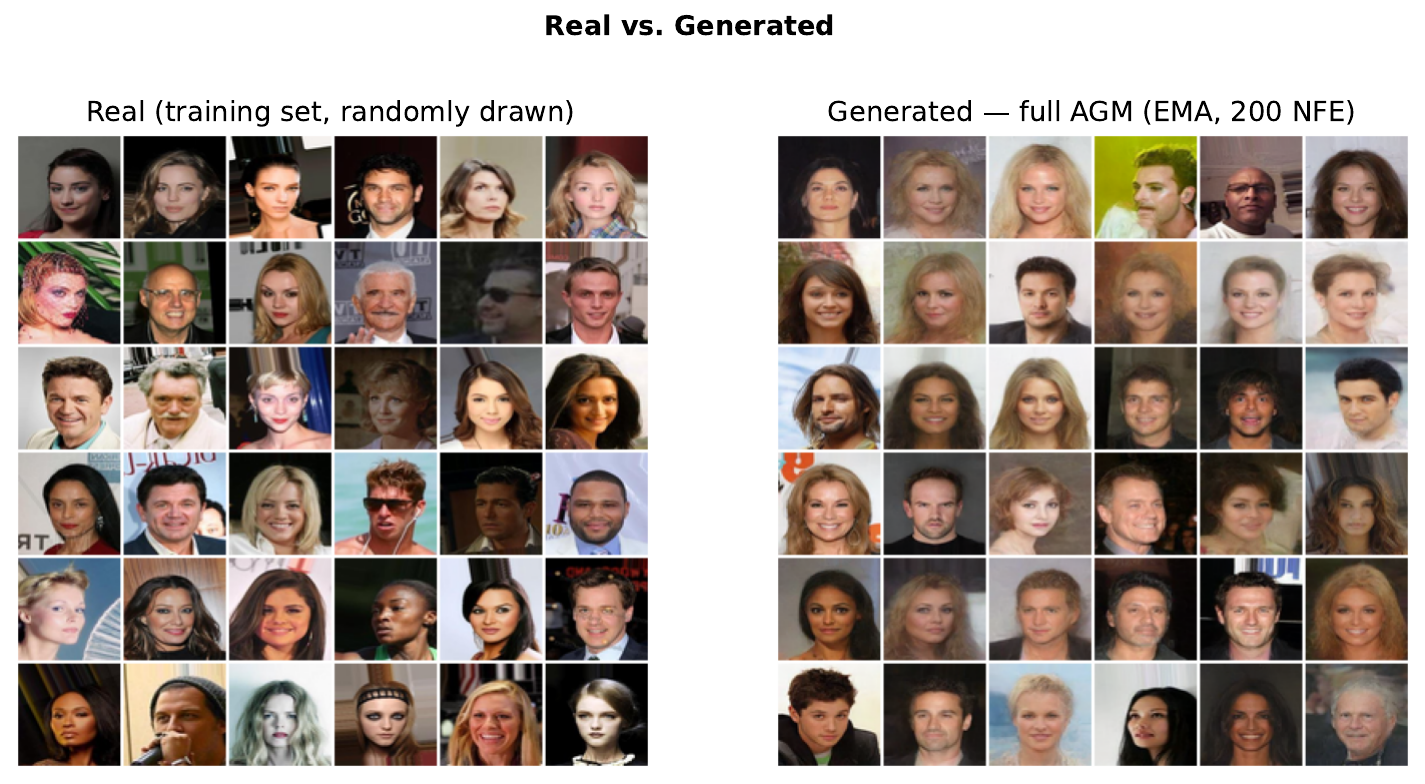}
  \caption{%
    \textbf{Real vs.\ generated.}
    Left: randomly drawn training images.
    Right: uncurated AGM samples (EMA drift, 200 NFE, no truncation).
    $V_\phi$ is not used at inference.%
  }
  \label{fig:rvsg}
\end{figure}

Figure~\ref{fig:rvsg} shows randomly drawn real images (left) alongside
uncurated AGM samples (right) generated with 200 NFE and no truncation.
The model produces coherent facial structure, realistic hair, and plausible
skin tones and backgrounds.  Minor artefacts in a small fraction of samples
(e.g., the unusual colouration visible in row 1, column 4) are expected at
this compute scale and dataset size.

\subsection{Training Dynamics}

\begin{figure}[htbp]
  \centering
  \includegraphics[width=0.62\linewidth]{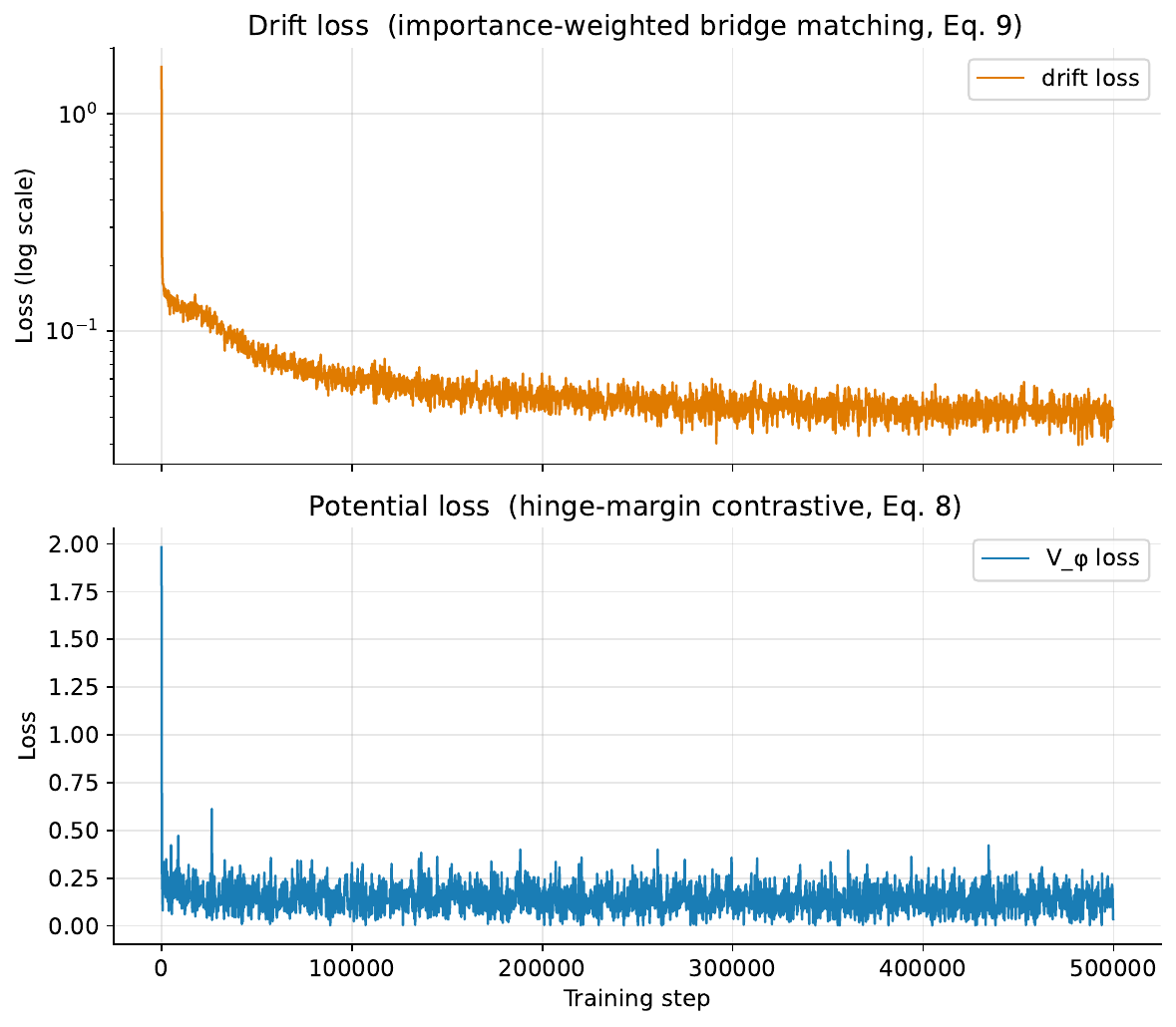}
  \caption{%
    \textbf{Training dynamics over 500{,}000 steps.}
    \textit{Top:} Drift loss $\mathcal{L}_f$ (importance-weighted bridge
    matching, Eq.~\ref{eq:loss_f}) on a logarithmic scale.
    \textit{Bottom:} Potential loss $\mathcal{L}_V$ (hinge-margin contrastive,
    Eq.~\ref{eq:loss_v}) on a linear scale.  Both losses stabilise
    cleanly, confirming that the two-network training scheme is well-behaved
    throughout.%
  }
  \label{fig:curves}
\end{figure}

Figure~\ref{fig:curves} shows the training curves over the full 500{,}000-step
run.  The upper panel shows the importance-weighted bridge-matching loss
$\mathcal{L}_f$ on a logarithmic scale: it drops sharply from $\sim$1.5 in
the first 10{,}000 steps and then continues to decrease steadily, stabilising
around 0.05 by step 200{,}000 with only minor oscillations thereafter.  This
smooth, monotone decline confirms that the drift network is learning
consistently and is not being destabilised by the importance weighting.

The lower panel shows the hinge-margin potential loss $\mathcal{L}_V$ on a
linear scale.  After a single large spike at initialisation (when $V_\phi$'s
outputs are uninitialised), $\mathcal{L}_V$ drops immediately to a low
plateau around 0.2 and remains there for the entire run.  The stability of
$\mathcal{L}_V$ confirms that the stop-gradient barrier is functioning as
intended: $V_\phi$ learns a stable separation between bridge samples and
noise without being drawn into adversarial oscillations with $f_\theta$.

Comparing the two loss curves directly reveals a further efficiency property:
the AGM drift loss descends more steeply during the first $\sim$50{,}000
steps than an equivalent drift-only run, indicating that $V_\phi$'s
importance weighting accelerates convergence rather than merely improving the
asymptote.  Checkpoint FIDs confirm this: the full AGM at approximately
350{,}000 steps matches or exceeds the FID achieved by the drift-only
baseline at 500{,}000 steps, representing a $\sim$30\% reduction in the
number of training steps required to reach equivalent generation quality.
This step efficiency is a practical benefit of path-aware weighting that
compounds the quality improvement: the same wall-clock budget delivers both
a better model \emph{and} a faster one.

\subsection{Potential Saliency: What Does $V_\phi$ Learn?}

\begin{figure}[htbp]
  \centering
  \includegraphics[width=0.80\linewidth]{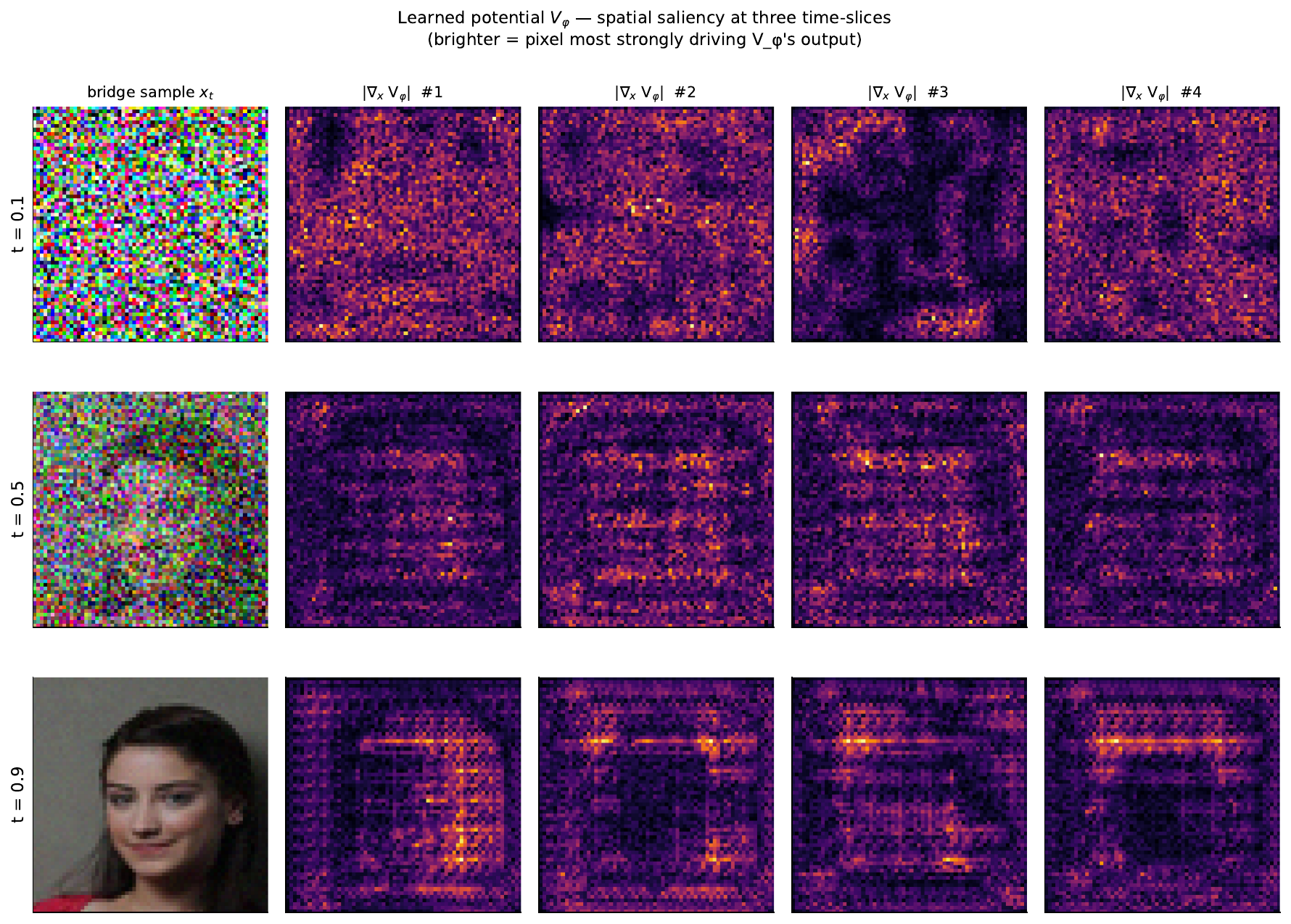}
  \caption{%
    \textbf{Learned potential $V_\phi$ --- spatial saliency.}
    Each row shows a bridge sample $\mathbf{x}_t$ (leftmost column) and
    the input-gradient magnitude $|\nabla_{\mathbf{x}} V_\phi|$ for four
    independently sampled bridges (brighter = higher saliency).
    At $t{=}0.1$ the signal is diffuse; at $t{=}0.9$ it localises precisely
    on facial structure, confirming that $V_\phi$ learns meaningful
    structural salience over the course of training.%
  }
  \label{fig:saliency}
\end{figure}

Figure~\ref{fig:saliency} visualises the spatial gradient of $V_\phi$ with
respect to the input bridge sample---i.e., the pixels that most strongly
drive the potential's output---at three time slices: $t=0.1$, $t=0.5$, and
$t=0.9$.

At $t=0.1$ (near-noise), $V_\phi$'s saliency is diffuse and roughly uniform
across the image, reflecting that all spatial regions carry similar potential
signal when the image is mostly Gaussian noise.  At $t=0.5$ (mid-bridge),
saliency concentrates in the regions where facial structure begins to emerge.
Strikingly, at $t=0.9$ (near-data), the saliency map sharpens dramatically
onto the face contour, hairline, and facial features---the exact structures
that define identity in the CelebA-HQ dataset.

This progression demonstrates that $V_\phi$ is not learning a trivial or
noise-driven signal.  It learns a time-aware notion of structural
informativeness: early in the bridge, all regions are equally uninformative,
but late in the bridge, the potential correctly identifies which spatial
details distinguish real faces from random noise.  The importance weights
derived from this signal therefore guide $f_\theta$ to invest more training
effort in learning the semantically meaningful parts of each bridge.

\FloatBarrier
\section{Conclusion}
\label{sec:conclusion}

We introduced \emph{Action-Inspired Generative Models} (AGMs), a dual-network
generative framework motivated by the observation that existing bridge-matching
methods treat all transport paths as equally informative.  By training a
lightweight scalar potential $V_\phi$ alongside the primary drift network
$f_\theta$---and coupling them exclusively through a stop-gradient importance
weighting mechanism---AGM re-focuses the drift's training signal on
structurally salient transitions while completely avoiding the adversarial
collapse that arises from unconstrained gradient flow between the two networks.

The core result is a 10.7\% improvement in FID (28.53 $\to$ 25.47), alongside
gains in both precision and recall, achieved over a drift-only baseline trained
under fully controlled conditions on a single NVIDIA RTX 4090.  The saliency
analysis of $V_\phi$ reveals that the potential learns semantically meaningful
spatial attention that sharpens progressively as the bridge nears the data end
of the trajectory---a qualitative validation that aligns with the mechanism's
theoretical motivation.

At inference, the framework reduces exactly to standard Euler--Maruyama
integration; $V_\phi$ adds zero overhead and is discarded entirely.  This
separation of training-time guidance from inference-time simplicity is, in our
view, one of the more appealing structural properties of the approach.

\paragraph{Limitations and future work.}
The current evaluation is limited to a single dataset and resolution, and we
want to be precise about why.  Scaling to $256{\times}256$ or above on a
single RTX 4090 is a genuine resource constraint: quadratic attention cost
and VRAM requirements for the DriftUNet at $C=128$ make 500{,}000-step runs
at that resolution infeasible within our budget, not a deliberate
deprioritisation.  Scientifically, however, higher resolutions are the
\emph{more interesting} regime for AGM: larger images have higher-dimensional
bridge manifolds and a correspondingly greater proportion of structurally
degenerate transitions, which is precisely the setting where
path-discriminating importance weighting is expected to be most impactful.
We therefore view resolution scaling as the highest-priority next step,
where the gains from $V_\phi$ are likely to be larger rather than smaller.
Training on more diverse and multimodal datasets is similarly important for
the same reason.  The connection to the principle of stationary action is
motivational rather than formal; establishing a rigorous variational
correspondence between $V_\phi$'s objective and an action functional for the
stochastic bridge would be a mathematically valuable extension.

\section*{Acknowledgements}
We gratefully acknowledge the Biomolecular Computation Laboratory,
Department of Computational and Data Sciences, Indian Institute of Science,
Bengaluru, for providing the GPU resources used in this work.

\bibliographystyle{unsrt}
\bibliography{references}

\end{document}